\documentclass[conference]{IEEEtran}
\IEEEoverridecommandlockouts
\usepackage{graphicx}
\usepackage{cite}
\usepackage{amsmath,amssymb,amsfonts}
\usepackage{algorithmic}
\usepackage{physics}
\usepackage{textcomp}
\usepackage{subcaption}
\usepackage[dvipsnames]{xcolor}
\usepackage{tabularx}
\usepackage[utf8x]{inputenc} 

\begin{document}

\title{Time and Cost-Efficient Bathymetric Mapping System using Sparse Point Cloud Generation and Automatic Object Detection\\
}

\author{\IEEEauthorblockN{Andres Pulido}
\IEEEauthorblockA{\textit{Mechanical and Aerospace Engineering} \\
\textit{University of Florida}\\
Gainesville, FL, United States \\
andrespulido@ufl.edu}
\and
\IEEEauthorblockN{Ruoyao Qin}
\IEEEauthorblockA{\textit{Mechanical and Aerospace Engineering} \\
\textit{University of Florida}\\
Gainesville, FL, United States \\
qinruoyao@ufl.edu}
\and
\IEEEauthorblockN{Antonio Diaz}
\IEEEauthorblockA{\textit{Mechanical and Aerospace Engineering} \\
\textit{University of Florida}\\
Gainesville, FL, United States \\
tony52892@ufl.edu}
\and
\IEEEauthorblockN{Andrew Ortega}
\IEEEauthorblockA{\textit{Geomatics Program} \\
\textit{University of Florida}\\
Gainesville, FL, United States \\
andrew.ortega@ufl.edu}
\and
\IEEEauthorblockN{Peter Ifju}
\IEEEauthorblockA{\textit{Mechanical and Aerospace Engineering} \\
\textit{University of Florida}\\
Gainesville, FL, United States \\
ifju@ufl.edu}
\and
\IEEEauthorblockN{Jaejeong (Jane) Shin}
\IEEEauthorblockA{\textit{Mechanical and Aerospace Engineering} \\
\textit{University of Florida}\\
Gainesville, FL, United States \\
jane.shin@ufl.edu}
}
\maketitle

\begin{abstract}
Generating 3D point cloud (PC) data from noisy sonar measurements is a problem that has potential applications for bathymetry mapping, artificial object inspection, mapping of aquatic plants and fauna as well as underwater navigation and localization of vehicles such as submarines. Side-scan sonar sensors are available in inexpensive cost ranges, especially in fish-finders, where the transducers are usually mounted to the bottom of a boat and can approach shallower depths than the ones attached to an Uncrewed Underwater Vehicle (UUV) can. However, extracting 3D information from side-scan sonar imagery is a difficult task because of its low signal-to-noise ratio and missing angle and depth information in the imagery. Since most algorithms that generate a 3D point cloud from side-scan sonar imagery use Shape from Shading (SFS) techniques, extracting 3D information is especially difficult when the seafloor is smooth, is slowly changing in depth, or does not have identifiable objects that make acoustic shadows. This paper introduces an efficient algorithm that generates a sparse 3D point cloud from side-scan sonar images. This computation is done in a computationally efficient manner by leveraging the geometry of the first sonar return combined with known positions provided by GPS and down-scan sonar depth measurement at each data point. Additionally, this paper implements another algorithm that uses a Convolutional Neural Network (CNN) using transfer learning to perform object detection on side-scan sonar images collected in real life and generated with a simulation. The algorithm was tested on both real and synthetic images to show reasonably accurate anomaly detection and classification. 
\end{abstract}


\section{Introduction}
Bathymetric maps take an important role in many applications by providing information of the underwater environment including water depth and geometry of the seafloor \cite{rs3010042,reef}. For example, in \cite{WEDDING20084159}, the authors obtained bathymetric mapping using LIDAR to determine nearshore benthic habitat complexity, which can assist ocean activity conservation and management. In \cite{SUB}, the authors introduces bathymetric and gravity technologies to reduce submarine fleets' dependence on GPS navigation. In \cite{LucidoTerrain}, the authors present a terrain-based underwater navigation using sonar bathymetric profiles. The natural change of seafloor can also be discovered by bathymetric mapping. An example work is shown in \cite{BathmetryChange}, where the authors used a bathymetric map to determine the relationship between natural change of seafloor to long-term shoreline change in century-scale. From this finding, shoreline change could be predicted by monitoring the change of seafloor, and this finding can help prevent erosion and reduce the risk of flooding. Bathymetric data also plays an important role in generating an ocean circulation model \cite{10.3389/fmars.2019.00283}, as the data can help researchers generate climate models and predict some global phenomena.

Although many applications require and benefit from bathymetric map data, obtaining such maps is often very time-consuming and costly. The most commonly-used sensor for bathymetric mapping is multi-beam echo-sounders, which are very costly because the sensors can send out multiple sound waves in a fan-shaped pattern at once and interpret the reflected signals. Moreover, multi-beam echo-sounders and other underwater sonar methods are generally done by a vessel operating on the surface of the sea \cite{10.3389/fmars.2019.00283}. One limitation of these methods is that the vessel must travel in a certain path which can only scan a limited region and also makes the job time-consuming. Another common way to obtain bathymetry in shallow clear water is to use lidar sensors with remote sensing technology. Lidar sensors transmit lasers from airborne platforms and measures their return \cite{10.3389/fmars.2019.00283}. This operation is usually performed using helicopters, satellites, and airplanes, which are not only expensive but also not able to map the sea floor with high resolution when the water is too deep or the water visibility is low.

Moreover, most existing bathymetric platforms focus on investigating a large scale area, such as ocean regions \cite{BathymetricDataViewerNOAA}. In order to provide quality bathymetry data for local reservoirs and retention ponds and overcome the aforementioned limitations, the authors in this paper developed a novel platform, named Bathy-drone, that can conduct bathymetric mapping for pond-scale bodies of water \cite{diazBathyDroneAutonomousUnmanned2022}. The proposed platform uses low-cost side-scan sonar sensors that are used in fish finders, which are generally available in affordable price ranges. Although the cost of these sensors is relatively lower than other methods, the sensors can still provide high-resolution bottom surface imagery \cite{rs13101945,rs13183555}. These type of sensors are proven to be useful in related applications, including marine life habitat tracking research \cite{NovelTechnique}. The Bathy-drone system is also designed to be lightweight such that a large transporting platform like a boat or an airplane is not required, and thus, the system requires reduced cost of operation. Autonomous or semi-autonomous systems can further reduce the number of people needed for operating the system and processing the sonar data. Furthermore, automation of the process and remote controlling of the system can access some region that is inaccessible or dangerous to operators \cite{10.3389/fmars.2019.00283}. 

The Bathy-drone platform requires an efficient and automatic mapping software, because the platform needs to be lightweight. The software system must have a very efficient sonar image processing algorithm given limited computation power and memory. Therefore, this paper proposes a computationally efficient algorithm that can map bathymetry using point cloud (PC). This paper presents our work on a bathymetric mapping system using sparse point cloud generation and automatic object detection with side-scan sonar. The proposed bathymetric mapping system consists of two algorithms. The first algorithm generates a sparse 3D point cloud data using side-scan sonar images, down-scan sonar depth readings, and GPS measurements from the sonar sensor. The second algorithm detects objects from the side-scan sonar images and outputs bounding boxes around the detected objects using YOLOv4 architecture and transfer learning \cite{bochkovskiy2020yolov4}. In order to overcome the lack of existing sonar images for training, we also generated some synthetic sonar data using Unreal Engine as well as real-life sonar data to perform fast and online side-scan sonar object detection. Real-life sonar data was acquired by the use of the Bathy-drone, a platform designed and manufactured by the University of Florida \cite{diazBathyDroneAutonomousUnmanned2022}.   

This paper is organized in four sections.  In Section \ref{sec:bathydrone}, the Bathy-drone platform is introduced with its specifications, such as the sensors equipped in the platform and the sensing resolutions. In Section \ref{sec:PCgeneration}, the efficient point cloud generation for bathymetric mapping is presented. The detail of the developed algorithms for object detection and sparse point cloud generation is included. In Section \ref{sec:experiment}, the field experiment results are presented. The authors performed the field experimentation at a retention pond at Citra, FL. The generated point cloud data of the pond is presented in this section.

\section{Bathy-drone: Time- and Cost-Efficient Bathymetric Mapping Platform}
\label{sec:bathydrone}
\begin{figure}[t]
    \centering
    \includegraphics[width=7.5cm]{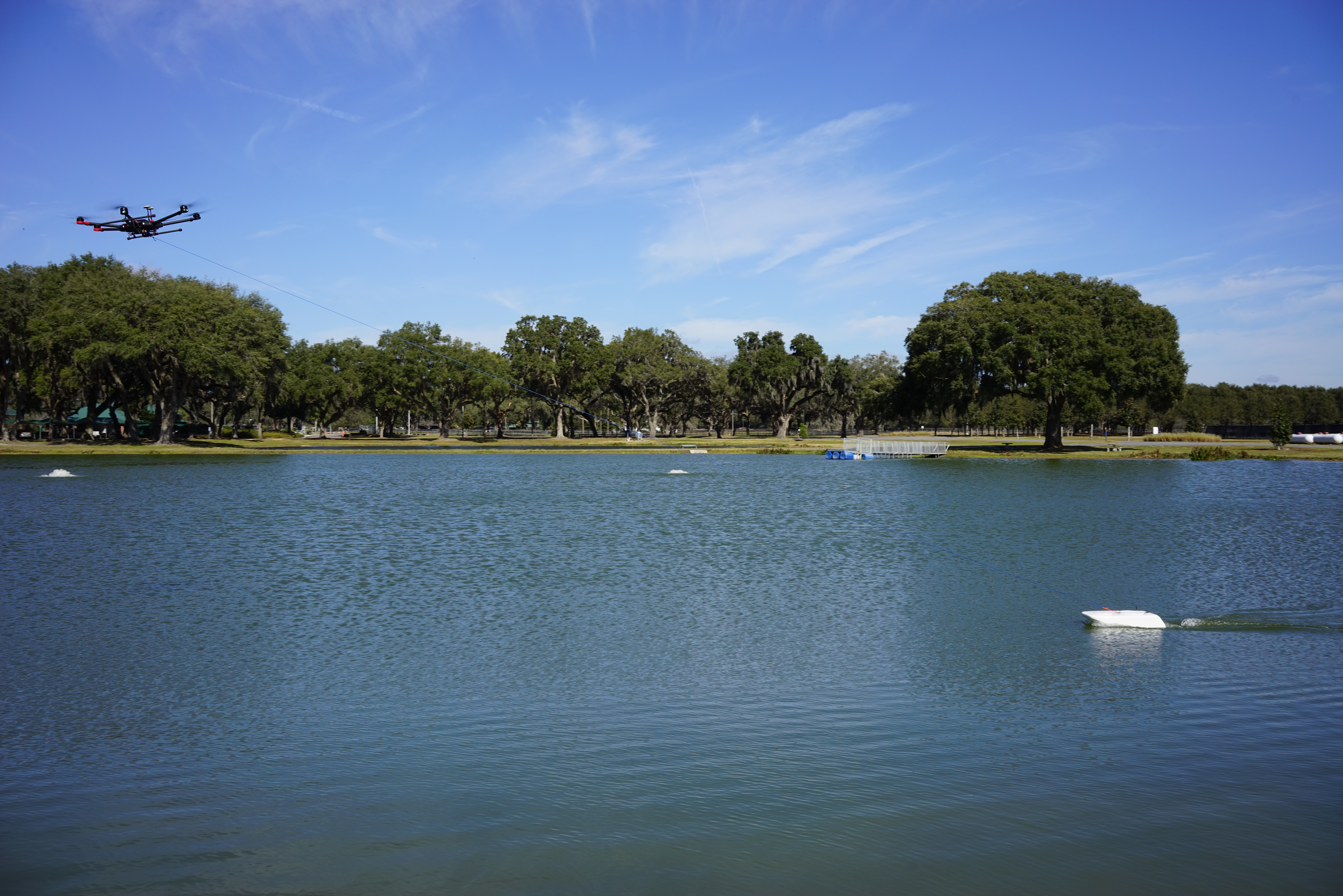}
    \caption{A photo of Bathy-drone system collecting data in a retention pond in Citra, FL.}
    \label{fig:bathydrone}
\end{figure}

The  Bathy-drone (Fig. \ref{fig:bathydrone}) is a semi-autonomous system that incorporates a drone towing a tethered vessel that can be equipped with a variety of sensors such as sonar or underwater cameras \cite{diazBathyDroneAutonomousUnmanned2022}. It has advantages over USVs since the system can be flown to the survey location; thus, many surveys can be initiated from a land-based ground station, and no boats or boat ramps are needed if the location is within the FAA's required visual line-of-sight. Additionally, since the vessel has no propulsion system (propellers) entanglement from flotsam, underwater or floating vegetation, does not hamper its operation. This allows it to navigate in small bodies of water, such as man-made and natural ponds, rivers, canals, low draft areas such as marshes and swamps, boat basins, shipping lanes, pre-construction and monitoring of marine infrastructure, and nearshore applications. The novel vehicle meets the industry and research need of developing specialized unmanned systems that are practical, inexpensive, easy to deploy and provide high spatial and temporal resolution for bathymetry and back-scatter. 

The Bathy-drone can collect bathymetry, sonar imagery, and bottom hardness while traveling at speeds of 0-24 km/h (0-15 mph) \cite{diazBathyDroneAutonomousUnmanned2022}. The tethered configuration and travel speed allow for use in waters with swift currents. An area of more than 10 acres was surveyed using the Bathy-drone on one drone battery charge and in less than 25 min\cite{diazBathyDroneAutonomousUnmanned2022}. The vessel is of fiberglass composite construction, with a soft-edge skiff-like planing hull shape. This shape proved robust in both straight/level tracking and corner turns once a trim plate and fins were added. Additional considerations in the design included providing adequate volume in the hull to house the sonar console and ports to provide easy access to the microSD cards, batteries, and sonar console keypad. The sonar unit is a low-cost commercial off-the-shelf recreational fish-finder by Lowrance, model Elite ti7, with an active scan transducer. This transducer provides side-scan and down-scan capability at multiple frequencies. The current Bathy-drone vessel and electronics payload weighs approximately 14 lbs. Heavy lift drones such as the DJI Matrice 600 used for this project can easily lift the vessel to the water surface. When testing for accuracy compared to the traditional state-of-the-art survey methods, the Bathy-drone produces contour plots where mean residual is −2.64 cm, the median is 0.95 cm, and the standard deviation is 16.98 cm\cite{diazBathyDroneAutonomousUnmanned2022}. When comparing repeated missions of the Bathy-drone for precision, the resulting residuals are summarized as a mean of 21.6 cm, a median of 18.7 cm, and standard deviation of 16.7 cm\cite{diazBathyDroneAutonomousUnmanned2022}. The resolution of bottom geometry is dependent on many factors such as shape, texture and density, but qualitatively objects as small as 0.5m (2-3ft) have been observed.


\section{Sparse Point Cloud Generation and Automatic Object Detection using Bathy-drone}
\label{sec:PCgeneration}
The down-scan depth and position data collected from the Bathy-drone's sonar scan can be used to generate underwater topographic maps called isobaths which are then converted to a point-cloud structure that can be helpful to understand the bottom geometry of the body of water. Point clouds can be used as a representation of bathymetry because they can be manipulated and stored with relatively low computational cost. However, only using the down-scan sonar depth only returns depth values at the path of the sonar. Therefore using the side-scan sonar images to generate additional points is useful to better represent the bathymetry. Additionally, during operation, detecting objects on the seabed using the side-scan sonar image also provides value to better understand the underwater environment. The same techniques to find the location of new points to the point cloud can be used to then locate the objects detected in the bathymetric map.

\subsection{Sparse Point Cloud Generation from Echo-sounder}
\begin{figure}
\centering
    \includegraphics[width=8cm]{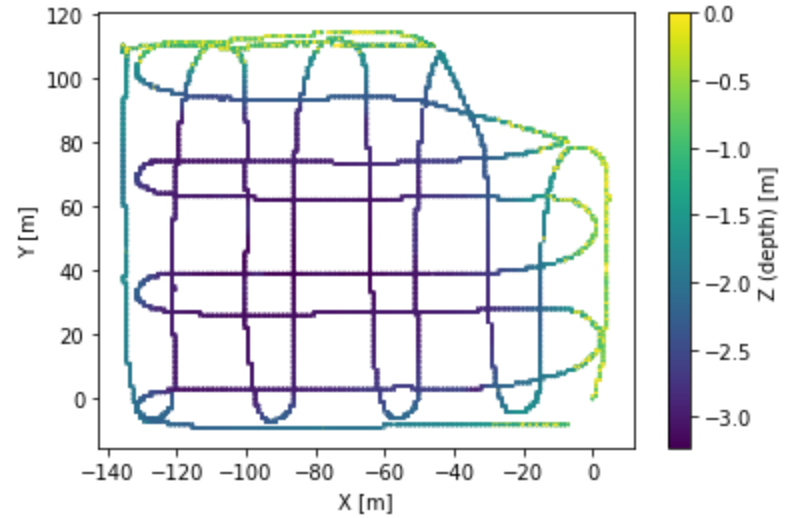}
    \caption{Path of the boat in ENU with the sampled depth from down-scan sonar}
    \label{fig:path}
\end{figure}

The algorithm presented in this paper generates a 3D point cloud in a computationally fast manner using only the side-scan sonar image with known position provided by GPS (the path of the boat, Fig. \ref{fig:path}), and down-scan sonar depth measurement at each data point. To accomplish this the side-scan image is assumed to have a property of linear mapping between distances to pixels on the port and starboard side. The output of the algorithm is an unordered point cloud data structure of $x,y,z$ points in the East North Up (ENU) coordinate system. 

For the Bathy-drone system, transformed GPS coordinates of latitude and longitude are used to find the planar $(x,y)$ position of the vehicle in time using the Spherical Pseudo-Mercator projection. The depth measurement provided by the down-scan gives the third coordinate to construct a sparse 3D PC. To increase the density of the PC, especially outside of the path traveled by the vehicle, the first return in side-scan sonar image is also used to generate additional points. The first return is the first reading of the water floor from each sonar beam. The first reading can be visually seen in a side-scan sonar image as the first brightly colored pixel after the `dead zone' in the middle of the image.

\begin{figure}
    \centering
    \includegraphics[width=8cm]{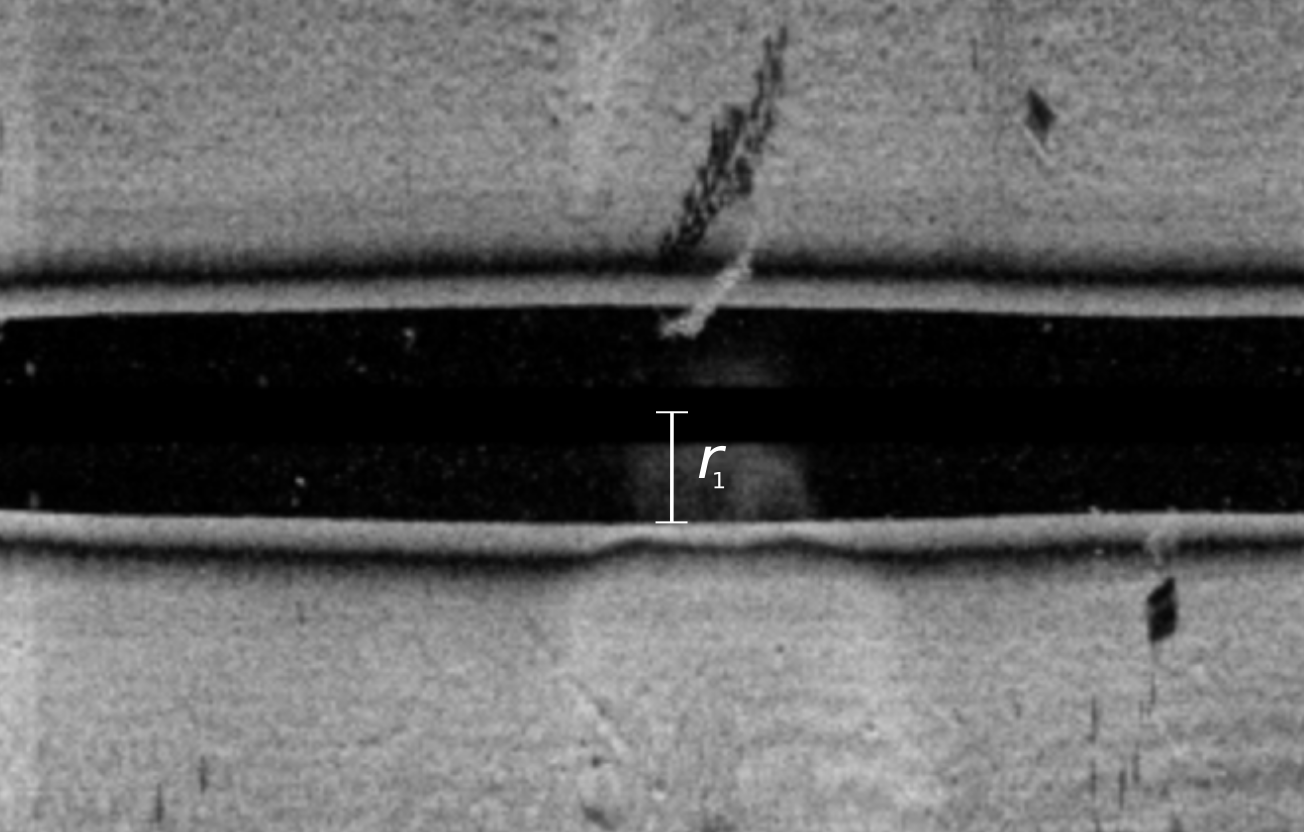}
    \caption{Pixel distance in the side-scan sonar image that is converted to $r_1$ with the $PPD$ ratio}
    \label{fig:sss_rad}
\end{figure}

\begin{figure}
  \centering
  \begin{subfigure}[t]{0.45\textwidth}
    \includegraphics[width=7.5cm]{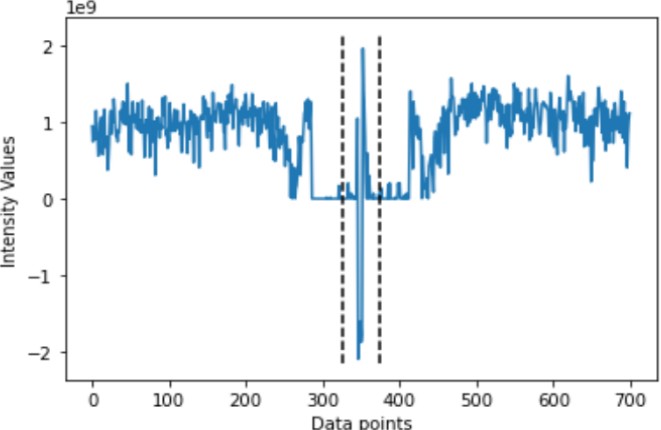}
    \caption{Raw intensity values of a side-scan sonar image column}
    \label{fig:scaling1}
  \end{subfigure}
  \begin{subfigure}[t]{0.45\textwidth}
    \includegraphics[width=7.5cm]{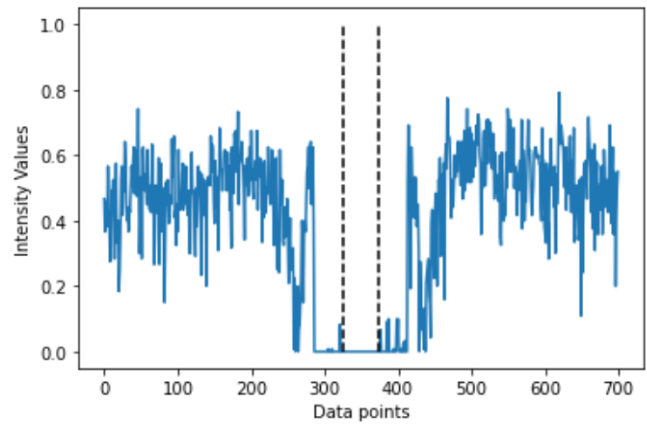}
    \caption{Scaled intensity values, and set the intensity values inside the black dots to zero}
    \label{fig:scaling2}
  \end{subfigure}
  \caption{Raw and normalized intensity values from the side-scan sonar}
\end{figure}

The first return of sonar reflection will be calculated using the sonar image (Fig. \ref{fig:sss_rad}). The raw image provided by the Lowrance fish-finder is a two-dimensional array of size $700 \times K$, where $K$ is a pixel number that grows as the sonar samples more points in a single image. The returned sonar intensity values are within the range of $I \in [0, 2.14] \times 10^{9}$ (Fig. \ref{fig:scaling1}). However, it is noted that the sonar image can have negative returned intensity near the middle of the deadzone (Fig. \ref{fig:sss_rad}) where the return is a pure reflection of the sonar itself. Therefore, the middle line values are set to zero in order to have only positive intensity. The intensity is scaled to be in between 0 and 1 in order to be able to generalize to other sonar sensors and have intuitive understanding of the intensity values as shown in Fig. \ref{fig:scaling2}. 

\begin{figure}
    \centering
    \includegraphics[width=8cm]{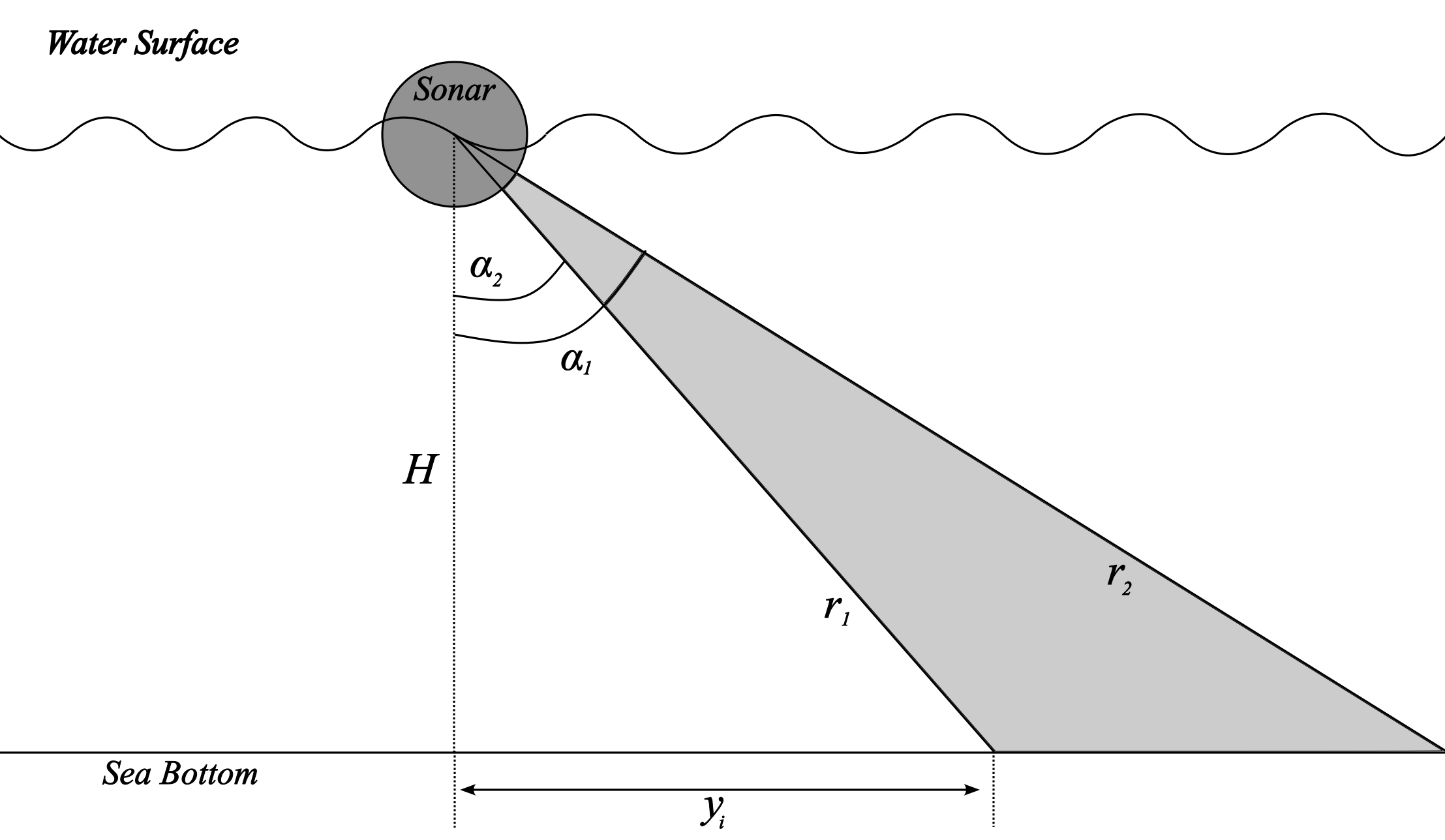}
    \caption{Geometry of the side-scan sonar beam used to get the first return}
    \label{fig:sss_angles}
\end{figure}

After the data is processed, the algorithm starts by using thresholding in the returned intensity to detect the first return. The algorithm finds the number of pixels from the middle of the image, the path of the robot, to the edge of the lighter-colored selection and maps those pixels to a distance in meters, $r_1$, in Fig. \ref{fig:sss_angles}). Using the assumed property of linear scaling between the pixels in the image and the distances in the image, a pixels per distance, denoted by $PPD$, is calculated by dividing $p$, the number of pixels for the first return, by $d$, the depth returned by down-scan sonar on a point in the map where the depth of the floor is known and the neighboring depth values are as close to flat as possible (\ref{ppd}). We determined this point by looking at the flattest portion of the pond in the depth-map interpolated using the down-scan sonar data. With the $PPD$ value, and the $\alpha_2$ angle found in the specification sheet provided by the sonar manufacturer, the depth at the first return on the starboard and the port side is calculated.

\begin{equation}\label{ppd}
    PPD = \frac{\text{(pixels)}}{\text{(distance)}} = \frac{p}{d} 
\end{equation}

\begin{figure}
    \centering
    \includegraphics[width=7cm]{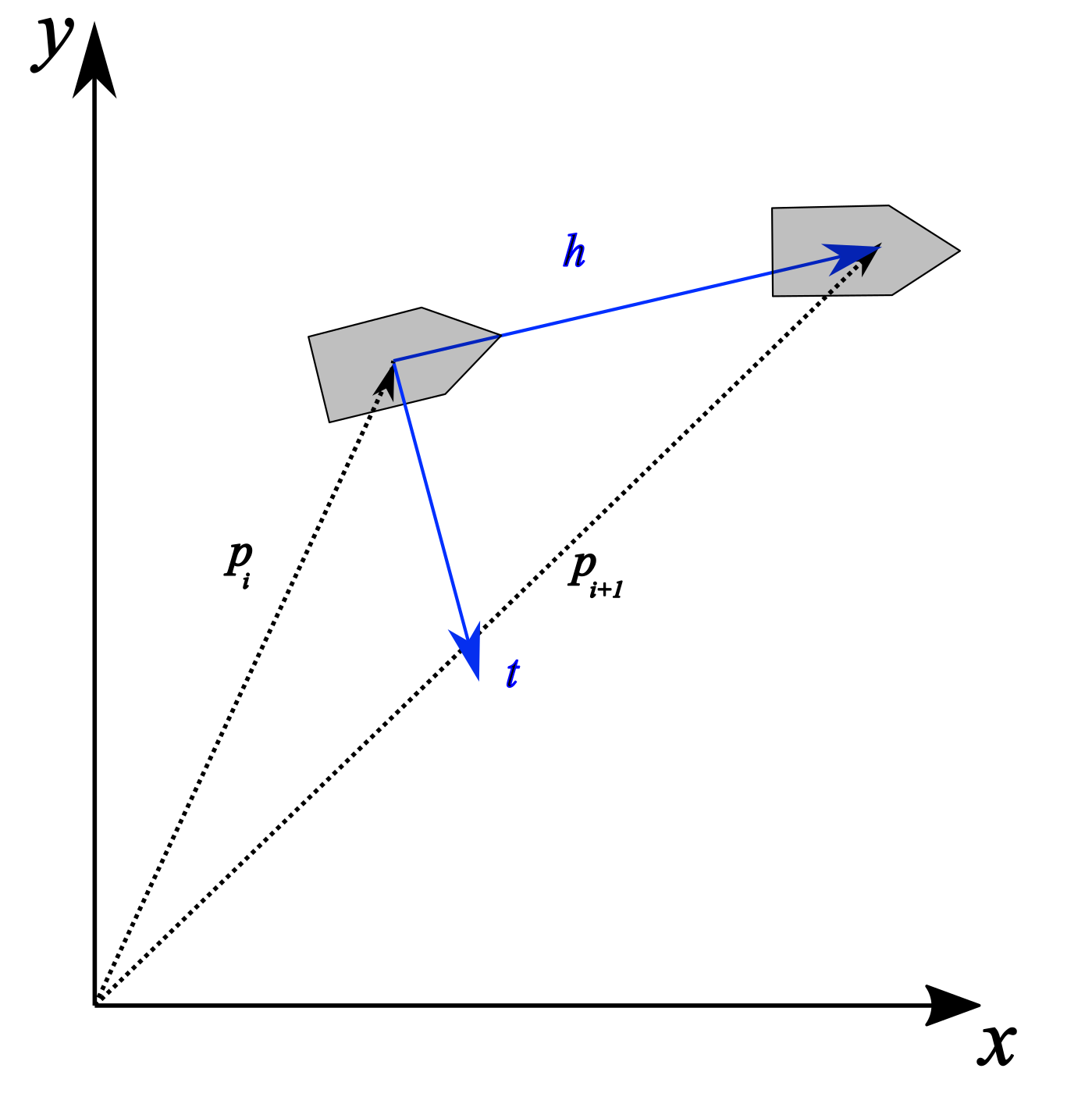}
    \caption{Heading and perpendicular vector geometries based on the boat position measurements}
    \label{fig:heading}
\end{figure}

The calculated depth is associated with a sampled point in the down-scan sonar, however, the point is in the reference frame of the boat, so it had to be transformed to the ENU coordinate system. To accomplish this, the heading \ref{heading} of the boat was calculated to be the direction of the $i$ sampled point to the $i+1$ sampled point,

\begin{equation}\label{heading}
    \vb{h_i} = \vb{p_{i+1}} - \vb{p_{i}} 
\end{equation}
\begin{equation}\label{perpendicular}
    \vb{t_i} = \vb{h_{i}} \cross (\vb{-\hat{z}}) \quad \textnormal{where} \quad \vb{\hat{z}} = [0,0,1] 
\end{equation}
\begin{equation}
    \vb{\hat{t_i}} = \frac{\vb{t_i}}{|\vb{t_i}|}
\end{equation}
Then the perpendicular vector \ref{perpendicular}, the direction where the side-scan sonar images are taken (Fig. \ref{fig:heading}), and the position of the boat at the sample point are used to project the distance in the starboard and the port side of the point to ENU using the distance $y_i$ \eqref{eq:horizontal}, seen in Fig. \ref{fig:sss_angles}, and found using trigonometry.
\begin{align}\label{eq:depth}
    z_i =& r_{i,1}\cos(\alpha_2) \\
    \label{eq:horizontal}
    y_i =& r_{i,1}\sin(\alpha_2) 
\end{align}
\begin{align} \label{eq:new_PC1}
    \vb{q_{(\text{starboard},\,i)}} =& \vb{p_i}+ \vb{\hat{t_i}}\,y_i  \\
    \label{eq:new_PC2}
    \vb{q_{(\text{port},\,i)}} =& \vb{p_i} - \vb{\hat{t_i}}\,y_i  
\end{align}

 The result is a 2D vector $\vb{q_{(j,\,i)}}$ \eqref{eq:new_PC1} and \eqref{eq:new_PC2} which is the position in the plane parallel to the surface of the water of the additional generated points from starboard and port side. To get the $(x,y,z)$ ordered list of the PC we append the $z_i$ depth \eqref{eq:depth}. 
 
 The final step of the process is to apply an outlier removal algorithm. The algorithm removes points that have few neighbors in a given sphere around them \cite{Zhou2018}. Two parameters can be changed to vary the minimum number of points inside the sphere and the radius of the sphere. 



\subsection{Automatic Object Detection Algorithm}
To perform the object detection task, the proposed work uses You-only-look-once(YOLO) structure. YOLOv4 is the fourth generation of the algorithm that uses neural networks to provide real-time object detection, which has the capability of detecting single or multiple objects in a single picture and predicting a bounding box around the objects. YOLOv4 consists of CSPDarknet53 as the backbone of the algorithm, spatial pyramid pooling (SPP) and Path Aggregation Network (PANet) as the neck, and YOLOv3 as the head. The SPP block over the CSPDarknet53 can increase the receptive field drastically and separate out the most significant context feature while having no negative impact on the operation speed. The PANet is the path-aggregation neck which is used as a method of parameter aggregation from different backbone levels for different detector levels \cite{bochkovskiy2020yolov4}. One of the reasons the YOLO architecture was chosen is that there has been previous work with previous versions (YOLOv2) to perform object detection using side-scan and forward-looking sonar images \cite{einsidlerDeepLearningApproach2018, karimanziraObjectDetectionSonar2020}.  
\begin{figure}[!htbp]
    \centering
    \includegraphics[width=8cm]{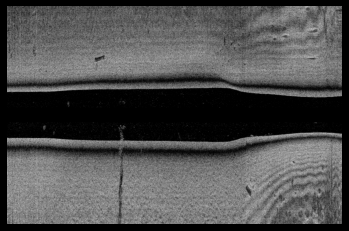}
    \caption{Example of side-Scan sonar image obtained from real underwater environment used for training}
    \label{fig:example1}
\end{figure}

\begin{figure}
     \centering
     \begin{subfigure}[b]{0.24\textwidth}
         \centering
         \includegraphics[width=\textwidth]{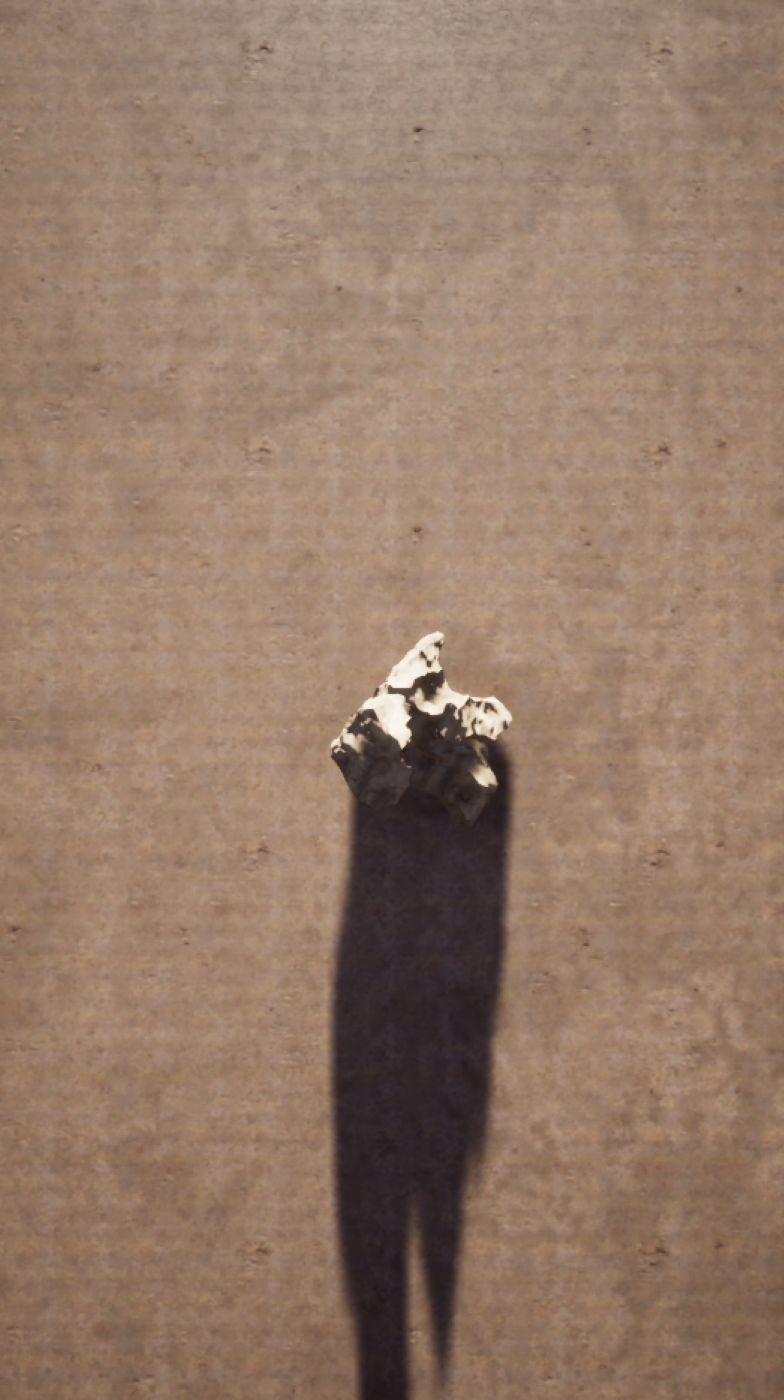}
         \caption{Example Image of an object obtained from Unreal Engine}
         \label{fig:example2}
     \end{subfigure}
     \begin{subfigure}[b]{0.24\textwidth}
         \centering
         \includegraphics[width=\textwidth]{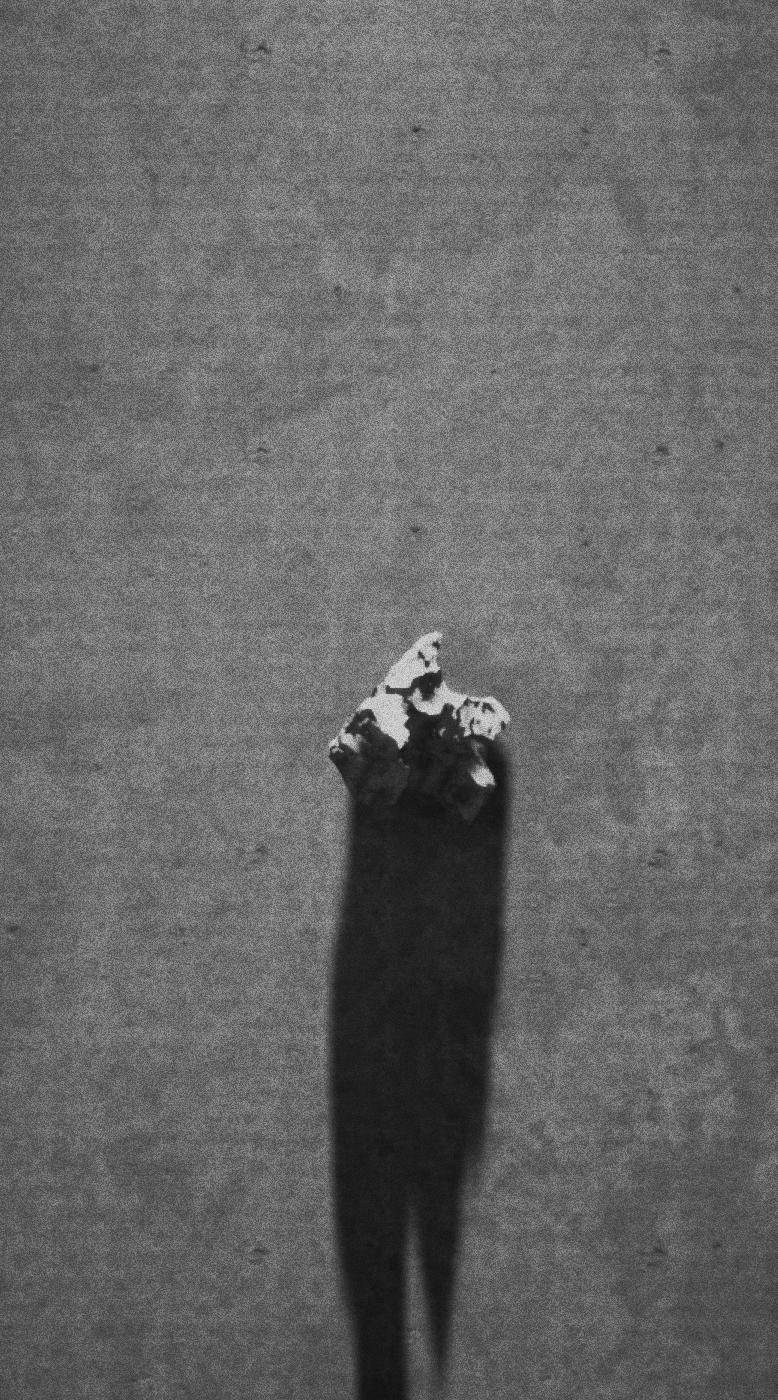}
         \caption{Image obtained from simulation after pre-processing}
         \label{fig:Simulation}
     \end{subfigure}
        \caption{Raw and pre-processed image generated in the UE simulator}
        \label{fig:two_ims}
\end{figure}

In the network training process, side-scan sonar images obtained from Bathy-drone platform (Fig. \ref{fig:example1}) and synthetic sonar images generated by a sonar image simulation are used. The synthetic data are used in the training process in order to obtain a batch of training images, which are not available from physical experiments due to time and cost constraints. The gaming framework, Unreal Engine (UE) is used as a simulation tool to generate synthetic sonar imagery data. The UE simulation models the way the sound waves get propagated underwater and then reflected by an object by approximating it as a light beam being generated from an overhead camera. An example of the generated synthetic images from the UE simulation is shown in Fig. \ref{fig:example2}. To increase the similarity between synthetic and real sonar images, a post-processing step was added. Specifically, the images are converted into grey scale, and then, speckle noise is added to the images. Fig. \ref{fig:Simulation} shows an example of the post-processed image.

Images obtained from UE after the pre-processing and real environment will form a data set that was divided into training, validation, and test set. There are 332 images in the training set, 41 images in the test set, and 33 images in the validation set. The performance of the YOLO network was evaluated with mean average precision (mAP) which is a common metric that evaluates the object detection model \cite{YOLO_metrics}. To calculate the mAP metric, average precision(AP) must be found for each class, and then, the average over the number of classes is taken.
\begin{equation}\label{eq:map}
    mAP = \frac{1}{N}\sum_{i=1}^{N}AP
\end{equation}
The Intersection over Union ($IoU$) is another metric that is used to evaluate the network \cite{Yu_2016_IoU,YOLO_metrics}. IoU represents the overlap of the predicted bounding box to the ground truth box.
\begin{equation}\label{eq:iou}
    IoU = \frac{\text{(Area of Overlap)}}{\text{(Area of Union)}}
\end{equation}
A higher IoU metric means the predicted bounding box is more similar to the ground truth bounding box. A confusion matrix is also used to determine the detection performance of the network.

\section{Experimental Results}
\label{sec:experiment}

\subsection{Sparse Point Cloud Generated from Echo-sounder}
\begin{figure}
    \centering
    \includegraphics[width=8cm]{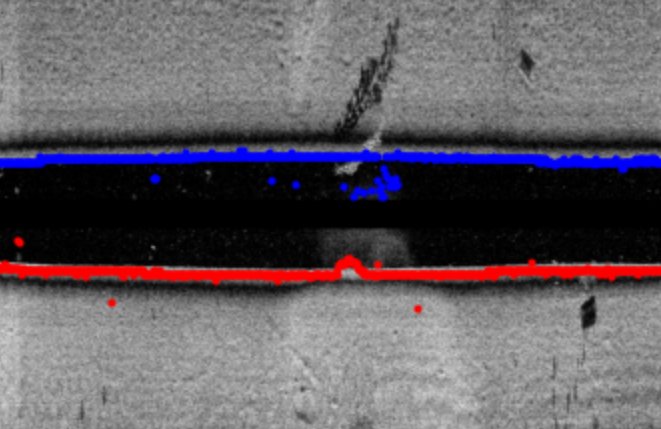}
    \caption{Side-scan sonar image strip with the first return colored red (starboard) and blue (port)}
    \label{fig:thresholding_strip}
\end{figure}

\begin{figure}
    \centering
    \includegraphics[width=8cm]{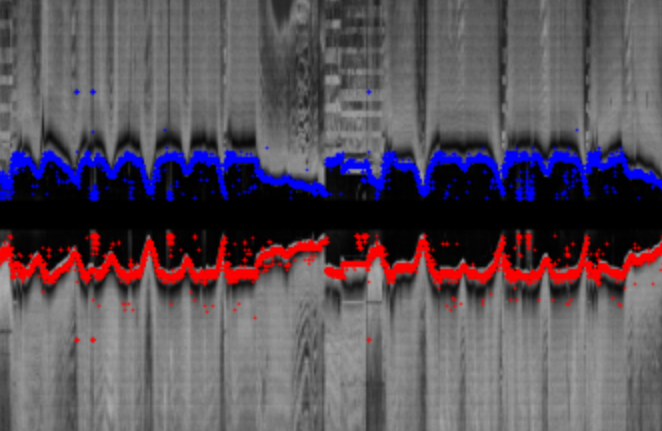}
    \caption{Side-scan sonar image of a complete run with the first return colored red (starboard) and blue (port)}
    \label{fig:thresholded_sss}
\end{figure}

The first return calculated by the algorithm of a side scan sonar image strip can be seen in Fig. \ref{fig:thresholding_strip}. The threshold value is set as 0.3. This threshold value needs to be between 0 and 1 because the side scan sonar image intensity values are scaled such that all values are ranged from 0 to 1. The first return for the entire side-scan sonar image of a run is shown in Fig. \ref{fig:thresholded_sss}.

\begin{figure}
    \centering
    \includegraphics[width=8cm]{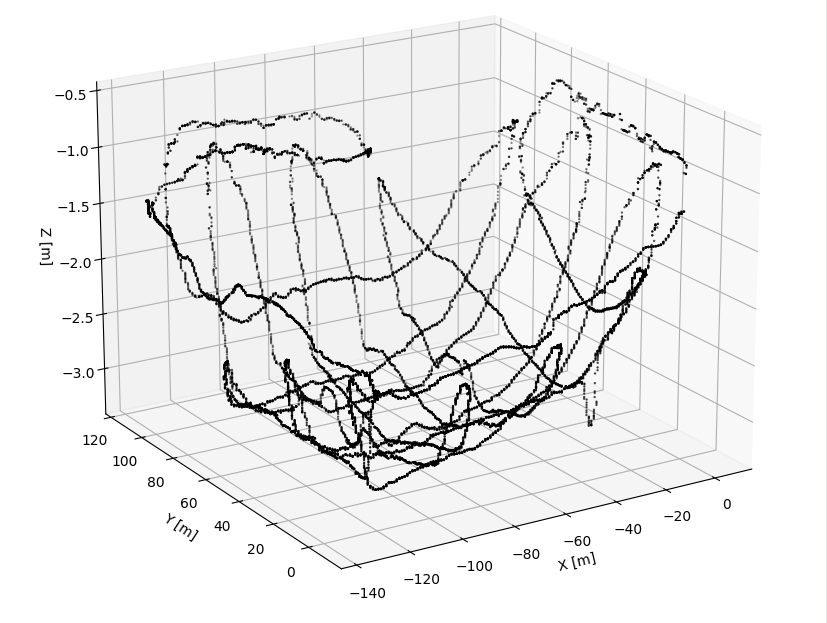}
    \caption{Point cloud data that are generated only using depth data from the path in Fig. \ref{fig:path}}
    \label{fig:downscan_PC}
\end{figure}

\begin{figure}
    \centering
    \includegraphics[width=8cm]{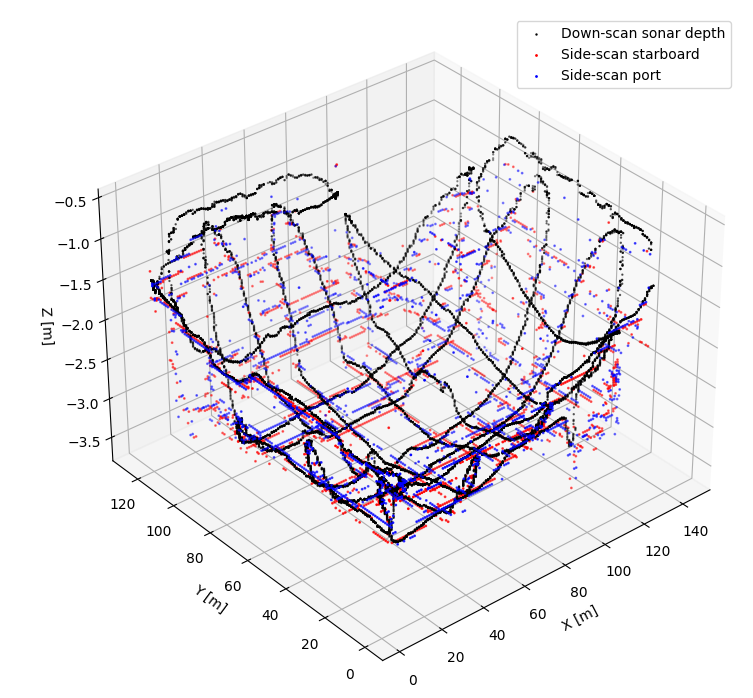}
    \caption{Sparse PC generated using the depth data and side-scan sonar images. The black points represent the points generated from the depth data, and the red and blue points represent the points generated from the starboard and port side of side-scan sonar images, respectively.}
    \label{fig:sparse_PC}
\end{figure}

The point cloud generated by only using the down-scan sonar can be seen in Fig. \ref{fig:downscan_PC}. The sparse PC generated using the algorithm explained above can be seen in Fig. \ref{fig:sparse_PC}, where the red and blue points are the ones calculated by the first return of the starboard and port sides respectively. The sparse point clouds can be seen to be less dense than the down-scan sonar cloud, a reason for this is that there are many points sampled at the same position and therefore the heading vector (and perpendicular vector) were found to be zero and no sparse point was calculated. Another reason for the sparsity is because approximately 10\% of the points are removed by the outlier removal algorithm \cite{Zhou2018}.  

\subsection{Automatic Object Detection Algorithm}
\begin{figure}
    \centering
    \includegraphics[width=4cm]{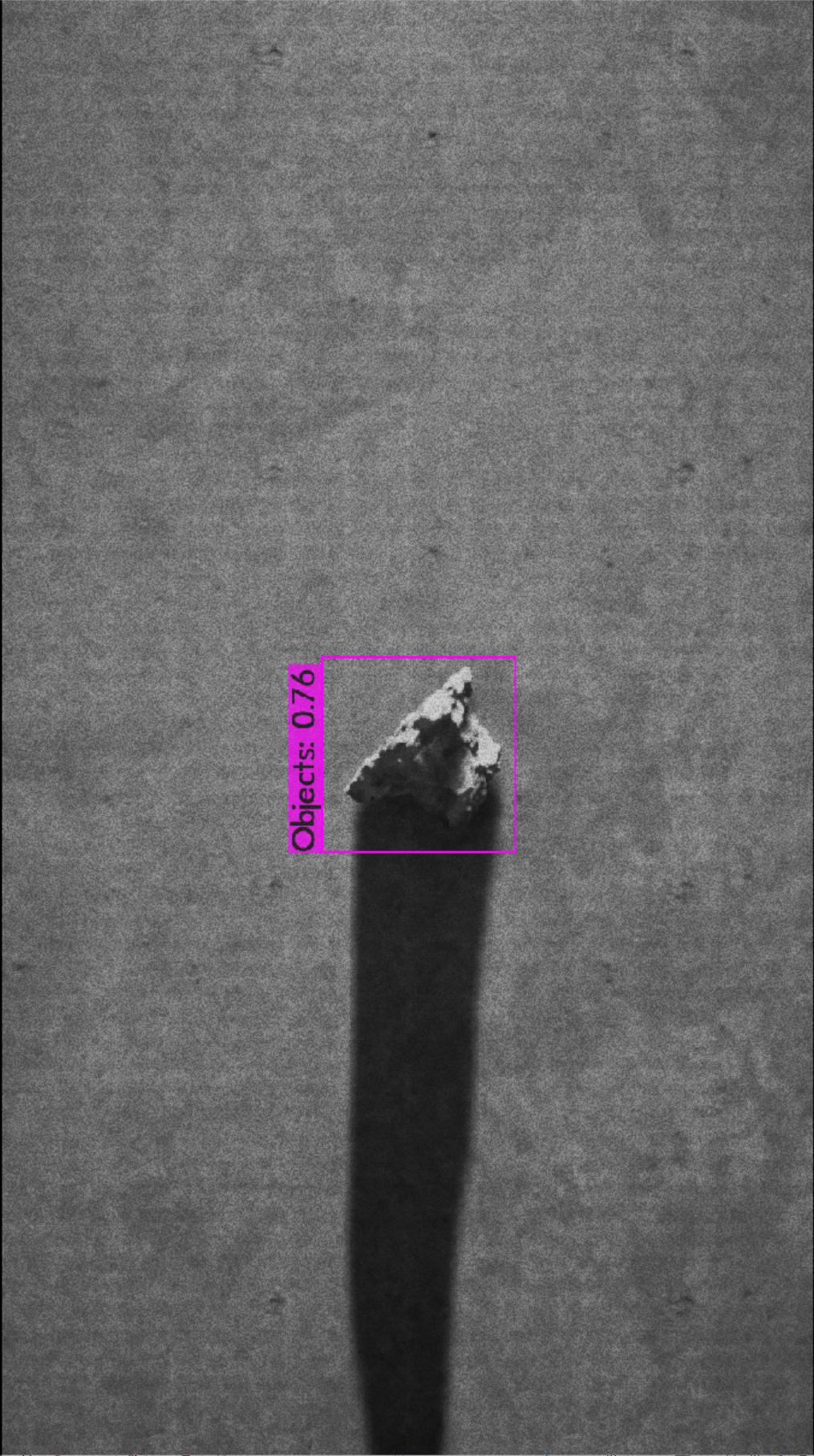}
    \caption{Objects detection on rock simulation image}
    \label{fig:Yolo}
\end{figure}

Fig. \ref{fig:Yolo} presents object detection results on a rock underwater that was generated from UE. 
Table \ref{metrics_table} presents the evaluation of the model with metrics of mAP and IoU and Table \ref{confusion_table} shows the confusion matrix values, in which True Positive (TP) means how many objects that the model predicts its positive class correctly. False Positive (FP) means how many objects the model predicts positive class incorrectly, and  False Negative (FN) means how many objects the model predicts the negative class incorrectly. 

\begin{figure}
    \centering
    \includegraphics[width=7cm]{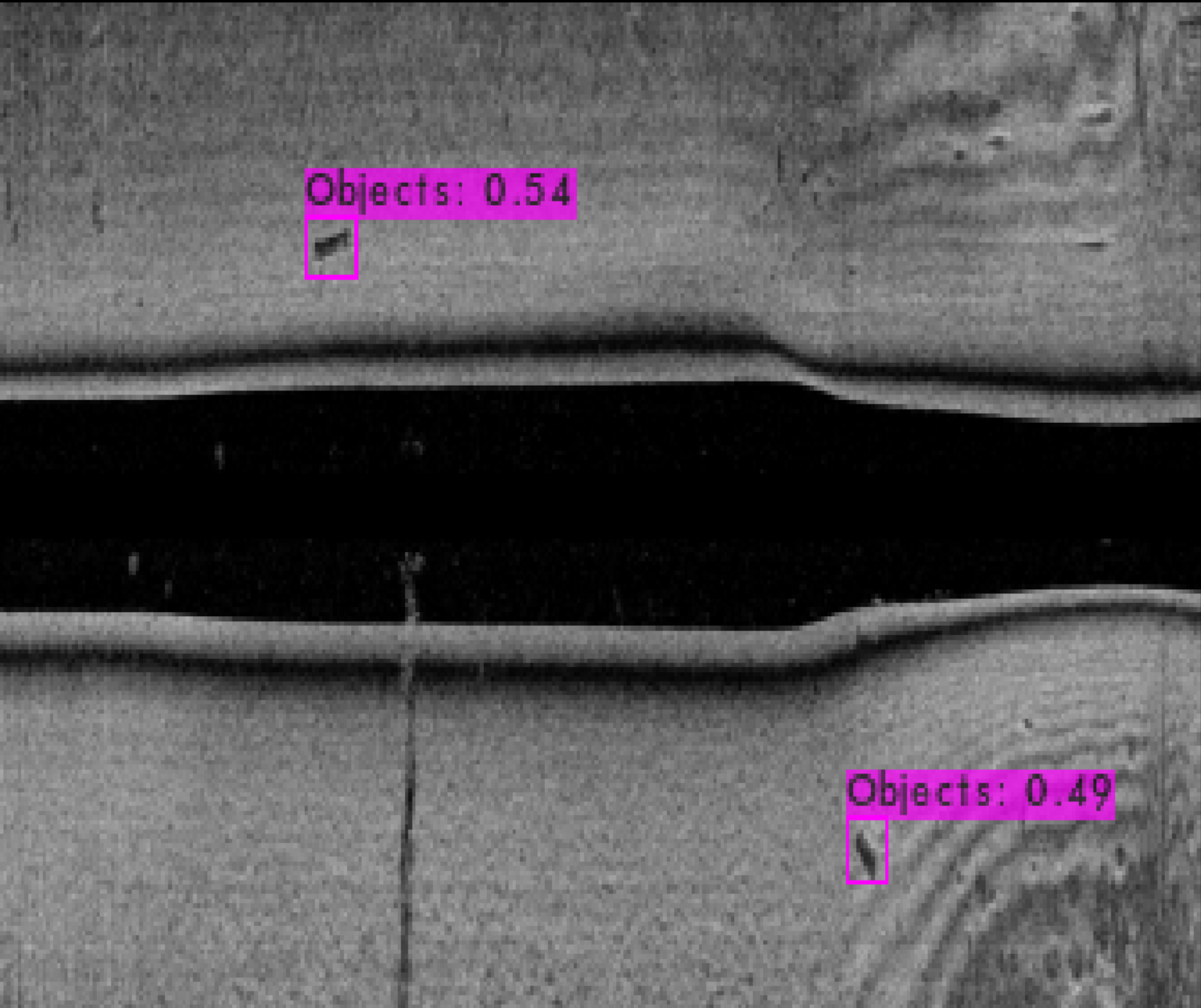}
    \caption{Object detection on in real Image}
    \label{fig:YOLO_Real}
\end{figure}

\begin{table}[h!]
\centering
\begin{tabular}{ccc}
\hline
         & Positive & Negative \\ \hline
Positive  & 34       & 46       \\ 
Negative  & 0        & 0        \\ \hline
\end{tabular}
\caption{Confusion matrix}
\label{confusion_table}
\end{table}

\begin{table}[h!] 
\centering
\begin{tabular}{cc}
\hline
Metrics & Value   \\ \hline
IoU     & 76.26\% \\ 
mAP     & 93.17\% \\ \hline
\end{tabular}
\caption{Metrics to evaluate the model}
\label{metrics_table}
\end{table}

Fig. \ref{fig:YOLO_Real} presents how the YOLOv4 algorithm performs object detection on the image obtained from UE. The number presented next to the box represents the object detection confidence level from the algorithm. The confidence level is high when the algorithm is tested using synthetic images as shown in Fig. \ref{fig:YOLO_Real}. However, the object detection confidence score drops on the images obtained from the real environment. The drop could be mitigated by collecting more actual sonar images of objects with the sides-scan sonar. 


\section{Conclusions}


This paper presents a computationally efficient point cloud generation algorithm for bathymetric mapping. The proposed algorithm is developed for mapping pond-scale water bodies using a low-cost platform named Bathy-drone. This Bathy-drone platform consists of a drone and an unpowered boat that is tethered to the drone. The boat includes a low-cost fish finder that includes down-scan and side-scan sonar sensors. The proposed algorithm generates a sparse 3D point cloud of the bathymetry using the down-scan and side-scan sonar measurement. The proposed algorithm first generates a sparse point cloud based on the depth data collected from the down-scan and the first returned signal of side-scan sonar sensors. Then, the algorithm performs automatic object detection from a side-scan sonar sensor in order to capture details in the pond. The proposed algorithm is tested based on the data obtained using Bathy-drone at a retention pond in Citra, FL. Future work will be dedicated to generate point cloud information of the objects detected using the side-scan sonar image. Shading techniques and a priori information of objects such as geometry from YOLO classification labels will be used to generate point clouds of the objects and include them in the bathymetric map. Based on the low computational complexity of the algorithm, another future work includes the real-time implementation of the algorithm onboard Bathy-drone such that the guidance system can automatically detect areas that need a closer look to complete the bathymetric map, and generate trajectories that the Bathy-drone will autonomously track. 

\section*{Acknowledgment}
The authors would like to thank the support from Aurigo Software Technologies, Inc.

\bibliographystyle{IEEEtran}
\bibliography{sparse_PC,YOLO}

\end{document}